\documentclass[letterpaper, 10 pt, conference]{ieeeconf}  

\IEEEoverridecommandlockouts                              

\usepackage{graphics} 
\usepackage[]{todonotes}
\usepackage[caption=false,font=footnotesize]{subfig}
\usepackage{multicol}
\usepackage{multirow}
\usepackage{amsmath}
\usepackage{amssymb}
\usepackage{comment}
\usepackage{booktabs}
\usepackage{cite} 
\usepackage[hidelinks]{hyperref}

\newcommand{\keywordsnew}[1]{\textbf{\textit{Keywords: }}{#1}}

\newcommand\copyrighttext{
	\footnotesize \textcopyright 2024 IEEE. Personal use of this material is permitted.  Permission from IEEE must be obtained for all other uses, in any current or future media, including reprinting/republishing this material for advertising or promotional purposes, creating new collective works, for resale or redistribution to servers or lists, or reuse of any copyrighted component of this work in other works. }
\newcommand\copyrightnotice{
	\begin{tikzpicture}[remember picture,overlay]
		\node[anchor=south,yshift=10pt] at (current page.south) {\fbox{\parbox{\dimexpr\textwidth-\fboxsep-\fboxrule\relax}{\copyrighttext}}};
	\end{tikzpicture}
}

\title{\LARGE \bf
Fast and Efficient Transformer-based Method for \\ Bird's Eye View Instance Prediction\\
}

\author{Miguel Antunes-Garc{\'i}a$^{1}$, Luis M. Bergasa$^{1}$, Santiago Montiel-Mar{\'i}n$^{1}$,\\ Rafael Barea$^{1}$, Fabio S{\'a}nchez-Garc{\'i}a$^{1}$, Angel Llamazares$^{1}$ 
\thanks{This work has been supported by the Spanish PID2021-126623OB-I00 project, funded by MICIN/AEI and FEDER, TED2021-130131A-I00, PDC2022-133470-I00 projects from MICIN/AEI and the European Union NextGenerationEU/PRTR, PLEC2023-010343 project (INARTRANS 4.0) from MCIN/AEI/10.13039/501100011033, and ELLIS Unit Madrid funded by Autonomous Community of Madrid.}
\thanks{$^{1}$M. Antunes-Garc{\'i}a, L.M. Bergasa, R. Barea, S. Montiel-Mar{\'i}n, F. S{\'a}nchez-Garc{\'i}a and A. Llamazares are with the Electronics Department, Universidad of Alcal{\'a} (UAH), Spain, {\{miguel.antunes, luism.bergasa, santiago.montiel, rafael.barea, fabio.sanchezg, angel.llamazares\}@uah.es}}
}

\begin{document}

\maketitle
\copyrightnotice
\thispagestyle{empty}
\pagestyle{empty}

\begin{abstract}


Accurate object detection and prediction are critical to ensure the safety and efficiency of self-driving architectures.
Predicting object trajectories and occupancy enables autonomous vehicles to anticipate movements and make decisions with future information, increasing their adaptability and reducing the risk of accidents.
Current State-Of-The-Art (SOTA) approaches often isolate the detection, tracking, and prediction stages, which can lead to significant prediction errors due to accumulated inaccuracies between stages.
Recent advances have improved the feature representation of multi-camera perception systems through Bird's-Eye View (BEV) transformations, boosting the development of end-to-end systems capable of predicting environmental elements directly from vehicle sensor data. These systems, however, often suffer from high processing times and number of parameters, creating challenges for real-world deployment.
To address these issues, this paper introduces a novel BEV instance prediction architecture based on a simplified paradigm that relies only on instance segmentation and flow prediction.
The proposed system prioritizes speed, aiming at reduced parameter counts and inference times compared to existing SOTA architectures, thanks to the incorporation of an efficient transformer-based architecture.
Furthermore, the implementation of the proposed architecture is optimized for performance improvements in PyTorch version 2.1.
Code and trained models are available at
\url{https://github.com/miguelag99/Efficient-Instance-Prediction}






\keywordsnew{Instance prediction, Autonomous Driving, NuScenes.}

\end{abstract}

\section{INTRODUCTION}


Nowadays, object detection, tracking, and prediction play critical roles in autonomous driving that have a direct impact on the safety and efficiency of Self Driving Vehicles (SDV).
Object detection and tracking give these systems the ability to identify and classify important environmental elements such as pedestrians and vehicles.
This is crucial to allow the vehicle to make the right decisions, avoiding collisions and granting the safety of the vehicle occupants, pedestrians, and other road users.
On the other hand, the prediction of object trajectories and occupancy allows SDV to anticipate movements and take proactive measures, improving their ability to adapt to dynamic situations and reducing the risk of accidents.
One State-Of-The-Art (SOTA) approach to this problem is to perform detection, tracking, and prediction independently so that the detection and tracking are focused on obtaining information about the past of the objects to perform the future prediction stage \cite{lanegcn},\cite{cratpred}. 
Using this approach in a real-world system can lead to significant errors in the prediction stage due to the accumulated errors between the stages.
It should also be noted that the performance evaluation of these motion prediction systems is usually based on the ground truth of the previous stages, resulting in metrics that do not consider the noise present in real systems, in which the detections and tracking are not perfect. 

\begin{figure}[t]
    \centering
    \includegraphics[width=0.45\textwidth]{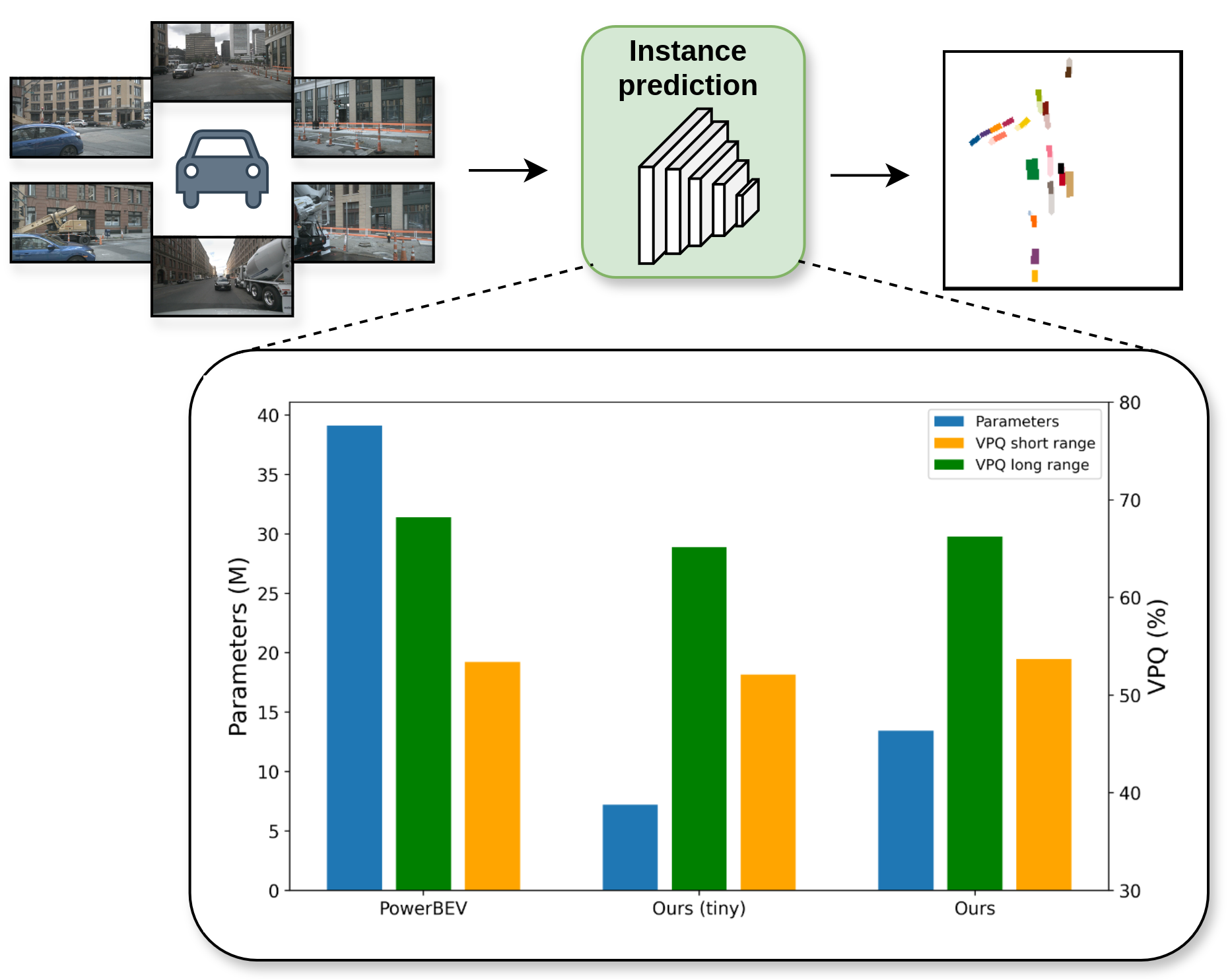}
    \caption{Our proposed architecture uses a multi-camera system to predict the position of the instances in the scene. We achieve similar results as our baseline PowerBEV reducing the number of parameters and latencies.}
    \vspace{-20pt}
    \label{fig:teaser}
\end{figure}


In recent years, techniques such as "Lift, Splat, Shoot" \cite{lss} have succeeded in improving the feature representation of multi-camera perception systems thanks to a Bird's-Eye View (BEV) transformation.
This representation maintains the three-dimensional and depth values of the vehicle's surroundings, discarding only the height information.
As a result, this has led to the evolution of end-to-end systems that directly predict environmental elements from the vehicle's sensor data.
Architectures such as \cite{fiery}\cite{zhang2022beverse} use the images provided by a multi-camera setup to generate information about the instances of the scenes as well as information about their future motion and occupancy.
Other SOTA approaches such as \cite{powerbev} propose a simplified instance prediction pipeline, training the system to perform only the BEV instance segmentation and the corresponding flow prediction.
Usually, these end-to-end systems have relatively high processing times and number of parameters, which can be a challenge if they have to be deployed in a real vehicle.

With this problem as the main focus, we propose a multi-camera BEV instance prediction architecture that uses the simplified paradigm presented in \cite{powerbev} and efficient attention modules specialized in dense tasks.
The proposed architecture aims for fewer parameters and inference time than other SOTA architectures.
We have also adapted the implementations of SOTA models to PyTorch version 2.1 for fair comparison.

\section{RELATED WORK}

\subsection{Camera 3D object detection}

The different image detection architectures have been evolving along with the appearance of new datasets focused on perception in autonomous driving.
KITTI \cite{kitti} set the foundation with its 3D detection benchmark, which only focuses on objects within the Field of View (FOV) of the front-facing stereo camera. 
Methods such as FCOS3D \cite{fcos3d} and PGD \cite{pgd} formulate the detection problem as a center-based multi-task learning of different 3D features such as size, rotation or centerness values.
On the other hand, methods such as SMOKE \cite{liu2020smoke} adopt a simpler approach, proposing the detection of the centers of the bounding boxes projected on the image instead of directly in 3D.
To improve the understanding of the scene, the different approaches have gradually focused on generating information from the 3D world instead of focusing on the image itself.
PseudoLidar++ \cite{you2020pseudolidar} obtains scene depth information from the images and represents it in a LiDAR-like point cloud, allowing a 3D LiDAR sensing architecture to be applied directly.

Detection architectures have evolved to handle complex multi-camera setups thanks to the release of autonomous driving datasets such as nuScenes \cite{nuscenes}.
This configuration offers a richer acquisition of \(360^o\) data from the environment, compared to a single-camera approach, which only acquires the portion in the frontal camera's Field-of-View (FOV).
DETR3D \cite{detr3d} uses object queries projected to 2D to obtain the corresponding features to generate object detections on a multi-camera system.
"Lift, Splat, Shoot" \cite{lss} proposes a unified representation of the features extracted from an arbitrary number of cameras using geometry and a latent depth distribution.
This method has significantly improved detection, tracking, and prediction over approaches that only use camera perspective representation of the 3D information.
BEVFormer \cite{li2022bevformer} performs the BEV transformation using an architecture that relies on self-attention and cross-attention to process the spatiotemporal information.

\subsection{Motion and Instance Prediction}

There are two major paradigms in SOTA regarding the prediction of the objects behavior in the vehicle environment: a) Motion prediction based on tracked objects. b) Instance prediction based on information from the sensors of the SDV.
In the first type, the architectures rely on detection and tracking systems that provide past information about the objects, which can introduce noise into the final prediction stage.
We can find architectures such as \cite{mercat_forecasting} or \cite{messaoud_attentionprediction}, which aim to perform the forecasting without extra information such as the scene map.
Both architectures use LSTM and multi-head self-attention to capture agent information.
CRATPred \cite{cratpred} improves the representation of social interactions by incorporating a GNN architecture and self-attention.
Other approaches such as \cite{lanegcn} or \cite{carlos_pred} extract information from the environment map to improve object predictions.

SOTA architectures have also been evolving towards approaches that do not rely on earlier stages of detection and tracking.
Fast and Furious \cite{luo2020fastandfurious} uses information from a single LiDAR sensor to generate detection, tracking and prediction information within a single network.
Fiery \cite{fiery} is one of the most influential methods, setting a milestone for instance prediction architectures.
Starting from a video with images from a system with six monocular cameras, the architecture generates a probabilistic prediction in BEV of the segmentation of the different instances and their respective future motions.
This approach is followed by other proposals such as BEVerse \cite{zhang2022beverse}, which uses a unified encoder with several specialized heads, or PowerBEV \cite{powerbev} with a simplified instance prediction approach based only on segmentation and flow.
These last three architectures incorporate a BEV projection of the extracted features from the images using the "Lift, Splat, Shoot" \cite{lss} approach, in which part of the latent space is dedicated to generating depth information before performing the projection.
On the other hand, TBP-Former \cite{Fang2023TBPFormerLT} proposes a PoseSync BEV Encoder block to transform camera features to BEV using cross-attention.

\subsection{Transformer-based architectures for dense tasks}


Attention-based architectures \cite{transformer} have brought a paradigm shift in Deep Learning since its introduction.
Although originally intended to be used with sequential data as in Natural Language Processing (NLP) tasks, their influence is also remarkable in areas of computer vision.
Vision Transformer (ViT) \cite{vit} is the precursor in this field, splitting the image into patches of the same size, which are transformed into the corresponding embeddings with the addition of another position embedding to maintain the spatial relations.
Swin Transformer \cite{liu2021swin} seeks to improve the representation of extracted features, combining the different patches in intermediate layers of the architecture, achieving better performance on large images.
Pyramid Vision Transformer (PVT) \cite{pvt} achieves an architecture based on attention mechanisms with a pyramid layout, similar to how it is implemented in multiple SOTA perception CNN architectures.
Instead of using the standard Multi-Head Attention (MHA) block, the authors propose Spatial-Reduction Attention (SRA), where the main difference is the reduction of the size of the keys (K) and values (V) before applying the MHA block.
SegFormer \cite{segformer} continues in this direction by proposing a multi-scale hierarchical encoder that does not require positional embeddings.
The features at different scales are merged in the architecture's final part, which helps achieve a lighter model.

\begin{figure*}[th]
    \centering
    \includegraphics[width=0.95\textwidth]{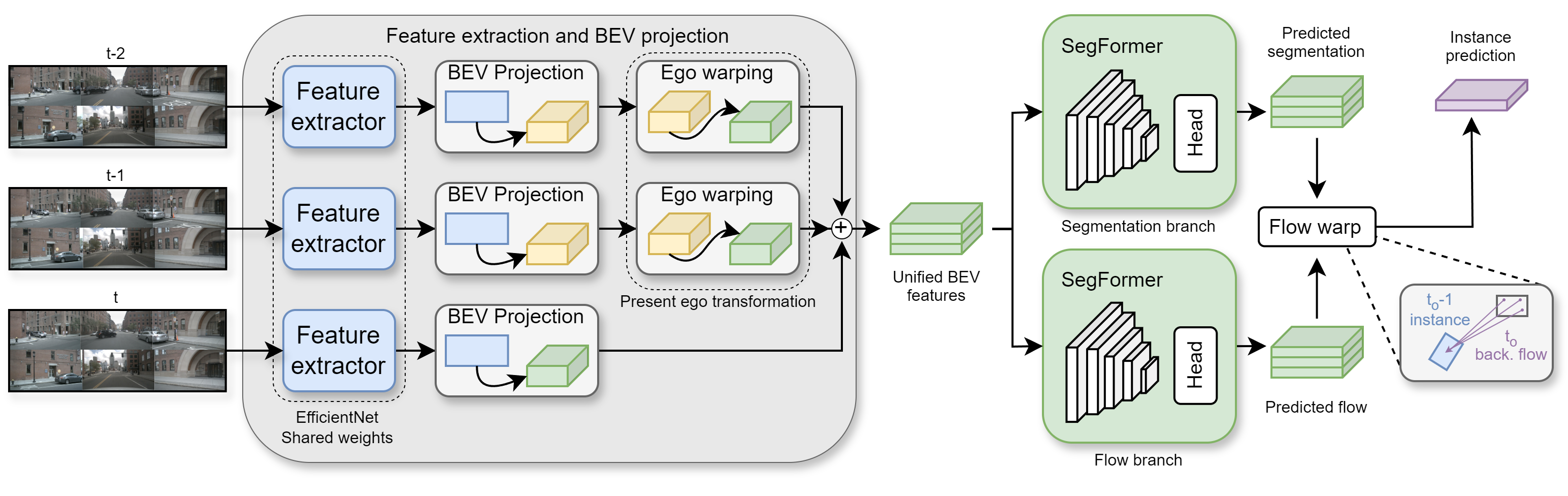}
    \caption{\textbf{Diagram of the proposed architecture}. First, the features of all the images are extracted for the whole input sequence. Each set of features of each instant is projected to the BEV using the information generated in the depth channels. For the past frames, it is necessary to apply a transformation that translates them to a unified system in the present frame. The generated BEV feature map is applied to two branches that generate the backward flow and segmentation values.}
    \label{fig:architecture}
\end{figure*}

\section{ARCHITECTURE}


Our proposed architecture is presented in detail in Fig. \ref{fig:architecture}.
The system uses the following input data from the cameras and GPS:
(a) The sweep of images taken from the multi-camera system \(I \in \mathbb{R}^{(T_p+1) \times N_c \times 3 \times H \times W}\).
(b) The intrinsic \(I_n \in \mathbb{R}^{N_c \times 3 \times 3}\) and extrinsic \(E_x \in \mathbb{R}^{N_c \times 4 \times 4}\) parameter matrices of each camera.
(c) The ego-vehicle location history \(P_{ego} \in \mathbb{R}^{(T_p+1) \times 7}\).
Where \(T_p\) is the number of past frames taken as input, \(N_c\) is the number of cameras in the vehicle setup and \(H\),\(W\) are the image dimensions.


The proposed architecture follows the paradigm proposed by PowerBEV \cite{powerbev}.
The model is tasked to generate only BEV maps of the instance segmentation  \(S \in \mathbb{R}^{T_f \times C \times H_{BEV} \times W_{BEV}}\) and backward flow values \(F \in \mathbb{R}^{T_f \times 2 \times H_{BEV} \times W_{BEV}}\) for each input sequence.
\(T_f\) is the number of frames to predict in the future, \(C\) is the number of output classes, and \((H_{BEV},W_{BEV})\) are the dimensions of the BEV map.
Note that the backward flow values define the displacement of each pixel of an instance in the current frame T to the center of the object in the previous frame T-1.

\subsection{Feature extraction and BEV projection}

The first stage of the network is responsible for extracting the features of the different cameras, generating the latent depth distribution, and making the corresponding projection to the BEV.
Like other instance prediction methods such as Fiery \cite{fiery} or PowerBEV \cite{powerbev}, we follow the approach proposed by "Lift, Splat, Shoot" \cite{lss} to achieve the unified representation of the environment features.
From each of the vehicle's cameras, we use \(T_p\) frames from the past plus the current frame, resulting in a total of \(Nc\cdot(T_p+1)\) frames for each input data sweep.
All of the images are processed by a single EfficientNet-B4 \cite{tan2020efficientnet} simultaneously, obtaining a feature map for each of them, in which there are \(C_D\) channels dedicated to depth information and \(C_F\) to the features of the environment itself.
Each of the extracted features with \(C_F\) channels is transformed into a BEV representation using an outer product with the information of the depth channels \(C_D\) generated by EfficientNet, obtaining a unified map in BEV \(F_{BEV} \in \mathbb{R}^{(T_p +1) \times C_{BEV} \times H_{BEV} \times W_{BEV}}\) with the features of all cameras for each instant.
The BEV map sizes (\(H_{BEV}, W_{BEV}\)) are defined by the resolution and distance limits that are configured for the X and Y axes (X and Z in the camera coordinate system).
Specifically, the distance covered on each axis is divided into equal parts, each corresponding to a particular position on the BEV map. 
To process all spatial information simultaneously, it is necessary to transform the features from past frames to the current frame of the vehicle.
Using the past positions of the ego-vehicle, we perform an ego-warping step where we project the past BEV features into the present by calculating and applying the transformation matrices between frames, achieving a unified representation aligned in the current ego while maintaining the dimensions and channels of the features.
Before applying the segmentation and flow branches, the information of the feature maps is reorganized, merging the temporal and spatial dimensions.

\subsection{Segmentation and flow branches}


Once we have managed to unify all the information from the cameras in a feature map \(F_{BEV}\), two parallel branches are incorporated to process all the spatio-temporal information, generating the segmentation and flow values.
The proposed architecture in \cite{powerbev} uses a U-Net based module in each branch to generate multi-scale information, which is crucial in such complex environments.
Our proposed model seeks to alleviate the computational impact introduced by the two branches, therefore we decide to implement an architecture based on SegFormer \cite{segformer} that efficiently uses attention to process the multi-scale features.
Five downsampling stages are incorporated in the encoder part, providing BEV features with sizes that are reduced by a factor of 2 in each one.
All encoder transformer blocks have a stride of 2, are composed of 2 layers, and have a hidden size 4 times larger than the input.
We propose two configurations of the architecture: a full version and a reduced version that helps to alleviate the computational load and the number of parameters even more.
The main difference between the two models is the channel configuration of the multi-scale features: $(16, 32, 64, 160, 256)$ for the full model and $(16, 24, 32, 48, 64)$ for the tiny version.



As in \cite{segformer}, each of the generated feature maps \(f_k, k \in \{1,...5\}\) is processed by an MLP + Upsampling block.
These upsampled maps are concatenated to produce a single unified map, which is used as the input to their respective head.

\begin{figure}[h]
    \centering
    \includegraphics[width=0.3\textwidth]{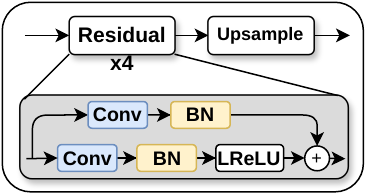}
    \caption{\textbf{Head architecture} for the segmentation and flow branches.}
    \label{fig:head}
    \vspace{-10pt}
\end{figure}

We propose the configuration shown in Fig.\ref{fig:head} as the head of each branch.
We incorporate four residual layers composed of a 2D convolution, batch normalization, and a Leaky ReLU activation function.
The number of channels is halved in the first and third residual layers, so a convolutional layer is added to adapt the residual connection.
The architecture is similar in both the segmentation and flow branches, changing only the dimensionality of the output maps of the last layer of the head.

\subsection{Flow warping}


To obtain the desired final representation of the different instances, it is necessary to propagate the information along the sequence.
As described in \cite{powerbev} and shown in Fig. \ref{fig:architecture}, in each frame of the input sequence, the instance values are updated with information from the previous frame and the corresponding backward flow.
Specifically, each BEV position in T is assigned with the ID of the instance of T-1 located at the destination of the generated flow.
With this approach, it is possible to propagate the instances at the individual level for each BEV position, minimizing potential errors in the association caused by failures in the prediction of the flow values.
There is no past information at instant \(T=0\), so an aggregation is performed to generate the IDs relying on the flow of the next frame.

\section{EXPERIMENTS}

\begin{figure*}
    \centering
    \includegraphics[width=0.9\textwidth]{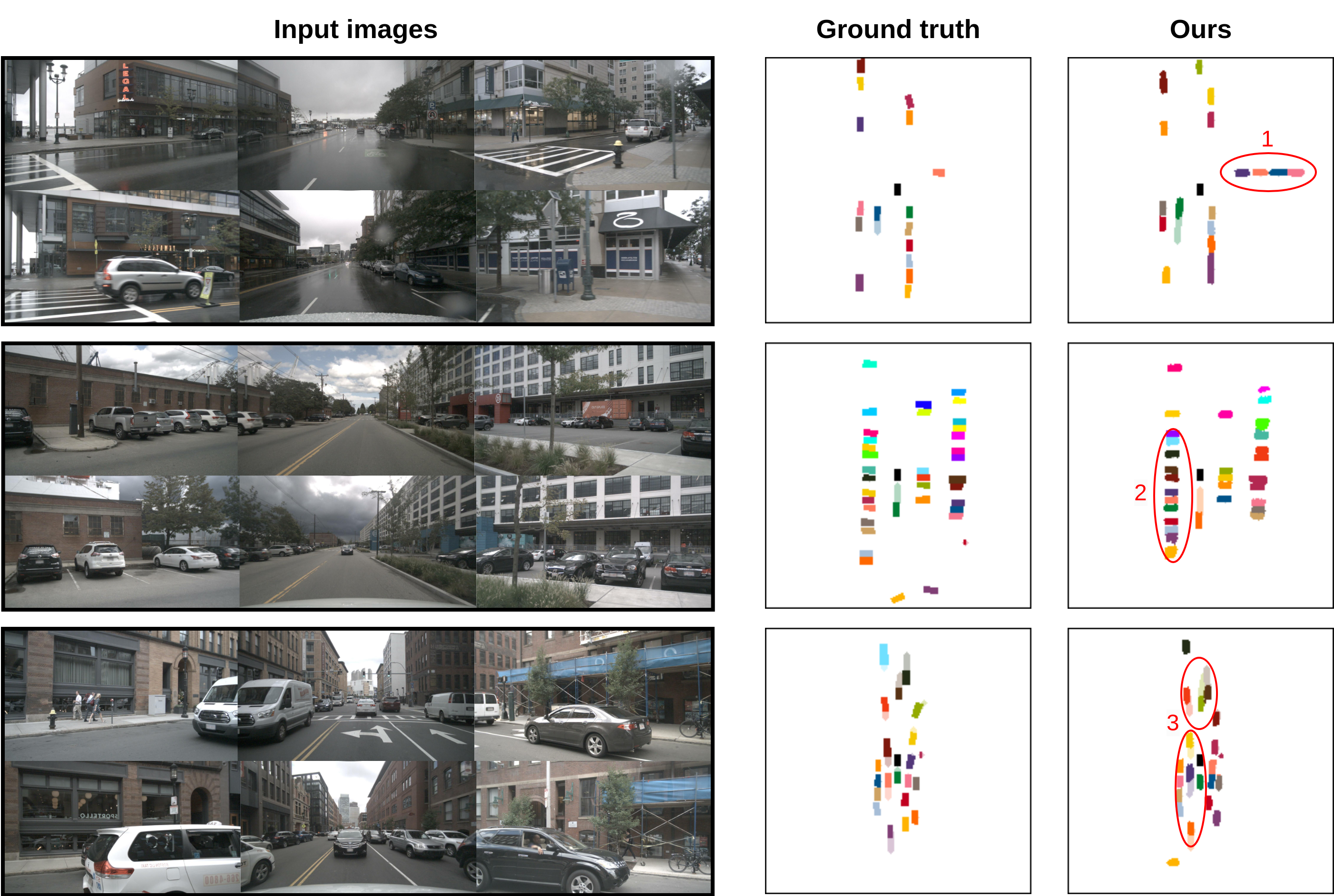}
    \caption{\textbf{Qualitative results on NuScenes validation set.} Each detected instance is represented by a color. If future positions are predicted, they are represented with the same color but with transparency. The ego vehicle is represented in black at the center.}
    \label{fig:qual_result}
\end{figure*}

\subsection{Implementation and Datasets}


The different experiments have been performed in the NuScenes \cite{nuscenes} autonomous driving dataset.
We use the official split of the 1000 scenes in the dataset to create 700 scenes for training, 150 for validation, and 150 for testing.
The data is collected at 2 Hz, with a total length of 20 seconds per scene.
As for the object classes, we generate the prediction ground truth values only with objects that belong to the \textit{'vehicle'} category.



The entire architecture and source code are implemented on PyTorch 2.1 and PyTorch Lightning.
The dataset images of size \(1600\times900\) are cropped and resized into a final input size of \(480\times224\) maintaining the same aspect ratio.
Regarding temporal information, we use values similar to the ones used in other state-of-the-art models to make a fair comparison.
Specifically, \(T_p=2\) is set, meaning that the network has information from 1 second in the past and from the current frame.
We also set \(T_f=4\), which results in a 2 seconds prediction into the future.
The experiments are performed with two spatial discretized BEV range configurations: an extended range of 100 meters (from -50 to 50) and a shorter range of 30 meters (from -15 to 15).
In both cases, the vehicle is in the center of the scene. 
In both the long and short configurations, we use grid resolutions of 50 cm and 15 cm, respectively. As a result, a fixed BEV grid size of 200 x 200 is achieved in the two scenarios.
To perform the 3D projection we need to define a distribution of depth values.
A distribution of 48 equidistant values from 2 to 50 meters is assigned. This configuration is used in the two defined BEV ranges.
This results in a value of \(C_D=48\) dedicated channels on the EfficientNet.

\subsection{Losses and metrics}

To perform the comparative performance study we used the same evaluation metrics as Fiery \cite{fiery} and PowerBEV \cite{powerbev}.
We first evaluate the ability of the architecture to perform the BEV segmentation of the vehicles in the scene.
We use the Intersection over Union (IoU) metric defined in Eq. \ref{eq:iou}, where \(\hat{y}_{t}^\text{seg}\) is the BEV map with the predicted segmentation and \({y}_{t}^\text{seg}\) is the scene segmentation ground-truth, both at the same timestamp \(t\).


\vspace{-10pt}
\begin{equation}
\operatorname{IoU}(\hat{y}_{t}^\text{seg}, {y}_{t}^\text{seg})=\frac{1}{T_\text{out}} \sum_{t=0}^{T_\text{out}-1} \frac{{\textstyle \sum_{h,w}} \hat{y}_{t}^\text{seg} \cdot {y}_{t}^\text{seg}}{{\textstyle \sum_{h,w} \hat{y}_{t}^\text{seg} + {y}_{t}^\text{seg} - \hat{y}_{t}^\text{seg} \cdot {y}_{t}^\text{seg}}},
\label{eq:iou}
\end{equation}


On the other hand, we evaluate the instance prediction over time using Video Panoptic Quality (VPQ), defined in Eq. \ref{eq:vpq}, where \(TP_t\) represents correctly detected instances, \(FP_t\) represents instances not associated with a ground-truth, and \(FN_t\) represents instances that have not been detected.

\begin{equation}
    \text{VPQ} = \sum_{t=0}^H \frac{\sum_{(p_t,q_t) \in TP_t} \text{IoU}(p_t,q_t)}{|TP_t| + \frac{1}{2}|FP_t| + \frac{1}{2}|FN_t|}
    \label{eq:vpq}
\end{equation}


Two losses, one for each output, are used to optimize the architecture.
For segmentation, we use a cross-entropy loss \(L_{seg}\) with \(k=25\%\), assigning importance to the worst predictions.
To optimize the flow, a loss \(L_{flow}\) of smooth-L1 distance is included.
The total loss (Eq. \ref{eq:loss}), is a combination of the two partial losses, with the addition of uncertainty weighting factors (\(\lambda_1,\lambda_2\)), which are updated during training.

\begin{equation}
	\mathcal{L} = \frac{1}{T_\text{f}} \left\{\sum_{t=0}^{T_\text{out}-1} \left(\lambda_1 \mathcal{L}_\text{seg} + \lambda_2 \mathcal{L}_\text{flow}\right)\right\}
\label{eq:loss}
\end{equation}

\subsection{Procedure and results}


We trained the different models for 20 epochs on the NuScenes train split.
The hardware used in all experiments was an Nvidia RTX 3090 GPU with 24 GB of VRAM.
As an optimizer, we used \textit{AdamW}, which was initialized with a Learning Rate (LR) of \(6e-5\) and gradually decreased according to a Polynomial scheduler.


First of all, we perform the comparison shown in Tab. \ref{tab:metrics}, which represents the metrics of different SOTA methods on the NuScenes validation split, using the two BEV range configurations.
All architectures are trained on the same setup so that the comparison is performed under the same conditions.
Given the great change in performance introduced by PyTorch versions higher than 2.0, we have updated the implementation of the studied models \cite{powerbev,fiery,StretchBEV,zhang2022beverse} to version 2.1, to perform the study of parameters and latencies.
All of these measurements represent the inference parameters of the models with a batch size set to 1, counting only the time of the architecture itself, without including the data preprocessing.
Note also that the values are obtained using the Python library Fvcore and the PyTorch interface.

\begin{table}[t]
\centering
\caption{Evaluation on NuScenes validation split.}
\label{tab:metrics}
\begin{tabular}{@{}ccccc@{}}
\toprule
\multirow{2}{*}{Method} & \multicolumn{2}{c}{VPQ $\uparrow$} & \multicolumn{2}{c}{IoU $\uparrow$} \\
                        & Short          & \;\;Long\;\;& Short       & \;\;Long\;\;\\ \midrule
\;\;\;\;PowerBEV \cite{powerbev}\;\;\;\;& 53.4          & 31.4         & 60.2           & \textbf{38.8}         \\
Fiery\cite{fiery}                        & 50.2          & 29.9         & 59.4           & 36.7         \\
StretchBEV\cite{StretchBEV}              & 46.0          & 29.0         & 55.5           & 37.1         \\
BEVerse \cite{zhang2022beverse}          & 52.2          & \textbf{33.3}& \textbf{60.3}  & 38.7         \\ \midrule
Ours                                     & \textbf{53.7} & 29.8         & 59.1           & 37.4         \\
Ours tiny                                & 52.3          & 28.9         & 57.5           & 36.9         \\ \bottomrule
\end{tabular}
\end{table}

\begin{table}[t]
\centering
\caption{Comparation of the different architectures on an RTX 3090 GPU using PyTorch 2.1 and Fvcore. Best in \textbf{bold}.}
\label{tab:params}
\begin{tabular}{@{}cccc@{}}
\toprule
Method     & Params (M) $\downarrow$      & VRAM (GB) $\downarrow$     & \;\;Lat. (ms) $\downarrow$\;\;          \\ \midrule
\;\;\;PowerBEV \cite{powerbev}\;\;\;   & 39.13          & 8.97          &  70               \\
Fiery\cite{fiery}      & 8.38  & 12.21         &  85               \\
StretchBEV\cite{StretchBEV} & 17.10          & \textbf{2.24}          &  900              \\
BEVerse \cite{zhang2022beverse}    & 102.5          & 7.84          & 
 230              \\ \midrule
Ours       & 13.46          & 8.19          &  63       \\
Ours tiny  & \textbf{7.42}  & 8.16          &  \textbf{60}      \\ \bottomrule
\end{tabular}
\vspace{-10pt}
\end{table}

The comparison with PowerBEV \cite{powerbev} is significant as it is our baseline and uses a similar approach.
As shown in  Tab. \ref{tab:metrics}, our model achieves remarkably close results at short distances, even outperforming it by 0.3 in VPQ, which leads to a more consistent instance identification through the sequence.
We measure a slight disadvantage of about one point in long-range scenarios.
Compared to the other SOTA models, our architecture achieves better VPQ at short range and is on par at long range, except for BEVerse, which improves at this distance setting.
This tendency is maintained for the IoU performance, achieving better results than Fiery and StretchBEV for the 50-meter configuration.

The comparison in Tab. \ref{tab:params} shows several efficiency parameters of the architectures, which are critical in autonomous driving.
It's important to note that our proposals are significantly lighter than other SOTA methods, with only 13 and 7 million parameters, values much lower than the 39 million of the baseline PowerBEV or the 17 and 100 million of StretchBEV and BEVerse respectively.
Furthermore, we also achieve a reduction in latency of about 10 ms compared to PowerBEV and 25 ms compared to Fiery.

Fig. \ref{fig:qual_result} shows some qualitative results, with 3 cases of interest: 1) The model successfully identifies and segments objects that were originally excluded from the ground truth due to poor visibility, showing its ability to generalization. 2) Faced with a scenario of high-density parked vehicles, the neural network manages to segment nearly all of them, indicating its effectiveness in detecting objects in crowded situations. 3) The model predicts the movement of both nearby and distant elements, as well as parked vehicles, even in scenarios with multiple instances, highlighting its ability to capture complex environment's context.
With all the information from the experiments, we can verify how our architecture offers similar results to other SOTA models but manages to cut inference times and the number of parameters with respect to its competitors.

\section{CONCLUSIONS AND FUTURE WORKS}

In this paper, we present a new architecture that uses efficient attention to perform instance prediction in multiple ranges.
Two configurations are proposed: normal and tiny.
A performance study is performed on the NuScenes dataset with several SOTA models, as well as a comparison of architecture efficiency parameters.
The results obtained allow us to confirm that despite being resource-efficient, both versions deliver instance prediction capabilities similar to state-of-the-art models.
In particular, our proposed architecture significantly lowers parameters and latency compared to other methods, making it a promising option for edge applications.
In the future, several possible improvements to the system are being considered.
First, the use of architectures with attention to generate BEV maps instead of using geometry with latent depth information.
Next, the use of information from other sensors to monitor the generation of the intermediate BEV map is proposed.
Finally, use the attention mechanism to better capture and process spatiotemporal information before applying the final two branches.

\addtolength{\textheight}{-12cm}   








\bibliographystyle{IEEEtranBST/IEEEtran}
\bibliography{bib}

\end{document}